\documentclass[conference]{IEEEtran}
\IEEEoverridecommandlockouts

\usepackage{cite}
\usepackage{amsmath,amssymb,amsfonts}
\usepackage{graphicx}
\usepackage{textcomp}
\usepackage{enumitem}
\usepackage{xcolor}
\usepackage{tikz}
\usepackage{verbatim}
\usetikzlibrary{shapes.gates.logic.US,trees,positioning,arrows}

\usepackage{algorithmicx}
\usepackage[ruled]{algorithm}
\usepackage{algpseudocode}

\usepackage{multirow}

\usepackage{caption}
\usepackage{subcaption}
\usepackage{changepage}
\usepackage{lipsum}
\usepackage[left=1.8cm,right=1.8cm,top=2.4cm,bottom=1.8cm,bindingoffset=0cm]{geometry}

\def\BibTeX{{\rm B\kern-.05em{\sc i\kern-.025em b}\kern-.08em
    T\kern-.1667em\lower.7ex\hbox{E}\kern-.125emX}}
\begin{document}

\title{
\huge Asynchronous Behavior Trees with Memory aimed at Aerial Vehicles with Redundancy in Flight Controller
\thanks{
$^1$The authors are with Space CREI, Skolkovo Institute of Science and
Technology (Skoltech), Moscow, Russia. \newline
$^2$The authors are with Institute of Robotics and Mechatronics, German
Aerospace Center (DLR), Wessling, Germany. e-mails: \newline  \texttt{evgenii.safronov@skoltech.ru, michael.vilzmann@dlr.de, d.tsetserukou@skoltech.ru, konstantin.kondak@dlr.de}
}
}

\author{\IEEEauthorblockN{Evgenii Safronov$^1$, Michael Vilzmann$^2$, Dzmitry Tsetserukou$^1$, and Konstantin Kondak$^2$}}

%

\newcommand{\mnt}{\emptyset}
\newcommand{\msR}{\mathbb{R}}
\newcommand{\msS}{\mathbb{S}}
\newcommand{\msF}{\mathbb{F}}

\newcommand{\nt}{$\emptyset$}
\newcommand{\sR}{$\mathbb{R}$}
\newcommand{\sS}{$\mathbb{S}$}
\newcommand{\sF}{$\mathbb{F}$}

\maketitle
\begin{abstract}
Complex aircraft systems are becoming a target for automation. For successful operation, they require both efficient and readable mission execution system (MES). Flight control computer (FCC) units, as well as all important subsystems, are often duplicated. Discrete nature of MES does not allow small differences in data flow among redundant FCCs which are acceptable for continuous control algorithms. Therefore, mission state consistency has to be specifically maintained. We present a novel MES which includes FCC state synchronization. To achieve this result we developed the new concept of Asynchronous Behavior Tree with Memory (ABTM) and proposed a state synchronization algorithm. The implemented system was tested and proven to work in a real-time simulation of High Altitude Pseudo Satellite (HAPS) mission.
\end{abstract}

\section{Introduction}
Nowadays the complexity of unmanned aerial vehicles (UAV) and their tasks is growing rapidly. Mission execution system interacts with all robot subsystems. Therefore, the size of the mission plan increases and it introduces the need for easy-readable, efficient, and reliable control architecture (CA) with FCC redundancy support. One of the common CAs is the Behavior Tree (BT), originated from game industry\cite{champandard2012behavior}. They have been widely used because of their modularity and reactivity properties. Behavior tree was suggested as a CA for unmanned aerial vehicles in\cite{ogren2012increasing}, \cite{klockner2013behavior}. There are several studies on BT applications in other areas of robotics, e.g. surgical robot\cite{surg} and robotic assistants\cite{ass}. Scientists touched many aspects of BT, such as a generalization of other CA\cite{generBT}, mathematical model formulation\cite{klockner2013interfacing}, automated BT construction \cite{LTL}, and others. The compherensive introduction to synchronous BT in robotics could be found in\cite{book}.
\begin{figure}[ht]
\centering
\includegraphics[width=0.48\textwidth]{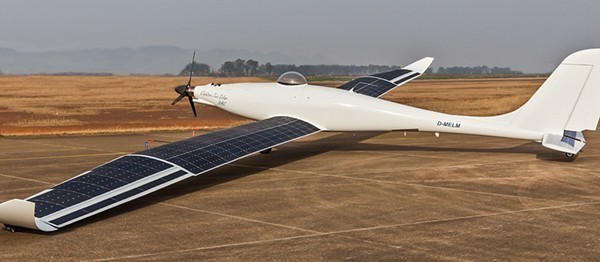}
\caption{Elektra Two Solar by Elektra Solar GmbH - an example of HAPS platform, one of the goal application for developed framework.}
\label{fig:haps}

\end{figure} 
\par One of the recent topics in this area is asynchronous (or event-driven) BT, which are deviant from classical synchronous BT formulation. It was shown that asynchronous BT outperforms synchronous in continuous-time mission simulation\cite{Klöckner}. Being flexible and reactive, behavior trees frequently require sort of inner memory, e.g., for a way-point missions. 

To solve this issue, the Reset and Latch node combination was suggested \cite{KlocknerStateful} as an alternative to control nodes with memory\cite{SwedishBT}.
A similar interaction between nodes could be  done by a ``blackboard" - a key-value table which is accessible by all tree nodes and sometimes other software modules. Although some authors argue that this approach ``does not lend itself well to encapsulation and, as a result, frustrates subtree reuse"\cite{parbt}, blackboards are widely used in both behavior tree frameworks for video game industry applications\cite{champandard2012behavior} and in robotics \cite{surg},\cite{ass}. 
\par
This work is devoted to designing a MES for UAV application. Our target platform is HAPS (Fig. \ref{fig:haps}). HAPS has to be autonomous to achieve continuous flight for several days or months. 
To achieve the required level of hardware reliability, HAPS is equipped with redundant flight control computers (FCC). However, this redundancy should be additionally supported on the software level.  Missing the data samples or different order of received messages would break  consistency of mission execution state on redundant FCC units. This issue might lead to unstable behavior of the HAPS, e.g., attempting execution of different commands. We developed statespace synchronization procedure to maintain state consistency on all FCCs. We took an advantage of \textit{asynchronous} BT propagation and formulated a new four tick propagation types to satisfy the goal for continous time simulation efficiency.
\par
In our BT architecture, interaction both with the other modules and between nodes is done completely by the blackboard, called hereafter \textit{memory}.
\par
We introduce a new Skipper control node which is symmetric to the well-described Selector and Sequence nodes.

\par 
The paper structured as follows: in Section II we describe changes in semantics and new nodes. Section III is devoted to explaining the asynchronous mechanism of call propagation in the tree from receiving a new information sample to sending out changes. In Section IV we describe how to support hardware redundancy on FCC with the developed framework. Section V denoted to implementation and conducted tests. In Section VI there are a conclusion and future work plans. 
\newgeometry{left=1.8cm,right=1.8cm,top=1.8cm,bottom=1.8cm,bindingoffset=0cm}

\section{Changes in Behavior Tree Formulation}
\paragraph{Semantics}
it is hard to name certain BT formulation as a canonical because almost every work on Behavior Tree slightly varies in semantics (see comparison\cite{SwedishBT}). The BT definition consists of a graph of nodes ($\mathcal{N}$) and edges ($\mathcal{E}$) which is a directed tree and a memory, which is a key-value dictionary ($\mathcal{V}$): 
\begin{equation}
t := \left\{\mathcal{G}\left(\mathcal{N}, \mathcal{E} \right), \mathcal{V} \right\}
\end{equation}
\par Each node in behavior tree has a state: $Running, Success$ or $Failure$. In the paper they are shorten respectively as \sR, \sS, \sF. Nodes are separated into two groups: leaf and control. Control nodes define the inner logic of the executor while leaf nodes are aimed to interact with the environment. Each node has a tick function, which updates its state due to the definition of the node.
\par Significant changes in our BT semantics comparing to other works are:
\begin{itemize}
\item Action and Condition nodes rely on external calls to corresponding middleware $\rightarrow$ Action and Condition nodes are functions over inner  memory.
\item Conditions nodes invoke binary (\sS, \sF) functions, Action nodes might be in (\sR)  state $\rightarrow$ Conditions might return one of (\sR, \sS, \sF)  states, Actions always return \sS.
\item The tick function is periodically applied to the root node, performing whole tree traversal $\rightarrow$ Only start of the execution is done by a single tick applied to the root node. The rest is executed in an asynchronous callback function.
\end{itemize}
\paragraph{Memory}
in our approach BT interacts with outer scope by assigning and reading the memory variables. For simplicity reasons, we assume that all variables have floating-point values and string keys. Each variable has a \textit{Scope} property: \begin{equation}
v \in \mathcal{V}, scope \left( v \right) \in \left\{
\begin{array}{l}
Input,\\ 
Output 
\end{array}
\right. 
\end{equation}
\par Only $Output$ variables would be added to return of callback if some Action changes their value. We also remember last result $state(\mathcal{N}) \subset \mathcal{V}$ of node evaluation. This is done to clearly define propagation rules. 
\paragraph{Leaf nodes}
as it is widely accepted \cite{champandard2012behavior},\cite{SwedishBT}, there are two leaf executable nodes which are able to interact with the outer scope (through $\mathcal{V}$): Action and Condition. In a memory-based tree, semantics of leaf nodes has to be changed. Action node modifies $\mathcal{V}$ and always returns \sS, as variables assignment could not fail or last for a significant time. Condition node does not modify the variables and returns a state $\in \left\{\msR, \msS, \msF\right\}$. On a change of at least one variable from this subset a condition should be reevaluated. 
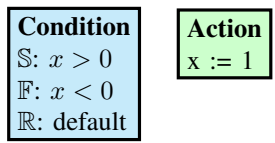
\begin{figure}[h]
    \centering
\begin{tikzpicture}[
        label distance=3mm,
        every label/.style={blue},
        event/.style={rectangle,thick,draw,fill=yellow!20,text width=2cm,
    		text centered,font=\sffamily,anchor=north},
        edge from parent/.style={very thick,draw=black!70},
        edge from parent path={(\tikzparentnode.south) -- ++(0,-1.05cm)
    			-| (\tikzchildnode.north)},
        level 1/.style={sibling distance=7cm,level distance=1.4cm,
    			growth parent anchor=south,nodes=event},
        level 2/.style={sibling distance=7cm},
        level 3/.style={sibling distance=6cm},
        level 4/.style={sibling distance=3cm}
        ]

	\begin{scope}[xshift=0cm,yshift=0cm,very thick,
		node distance=1.6cm,on grid,>=stealth',
		cond/.style={rectangle,draw,fill=cyan!20},
		act/.style={rectangle,draw,fill=green!20},
		comp_g/.style={circle,draw,fill=green!20},
		comp_d/.style={circle,draw,fill=red!20},
		comp/.style={circle,draw,fill=orange!40}]
   
   \node [cond] (C)	[xshift=4cm, yshift=-0.4cm, align=left]{\textbf{Condition} \\ \sS: $x > 0$  \\ \sF: $x < 0$ \\ \sR: default} ;
   \node [act] (A)  [xshift=6cm, align=left]{\textbf{Action} \\ x := 1};
   \end{scope}
\end{tikzpicture}
\caption{Condition and Action nodes example.}
    \label{fig:condact}
\end{figure}
\par
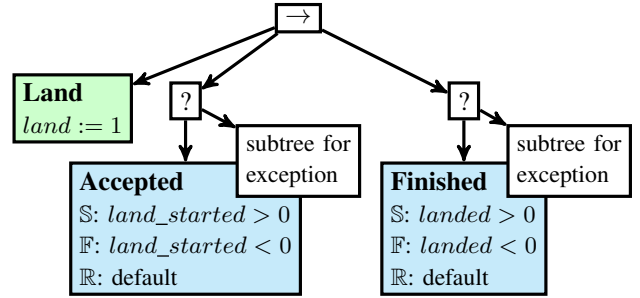
\begin{figure}[h]
    \centering
\begin{tikzpicture}[
        label distance=3mm,
        every label/.style={blue},
        event/.style={rectangle,thick,draw,fill=yellow!20,text width=2cm,
    		text centered,font=\sffamily,anchor=north},
        edge from parent/.style={very thick,draw=black!70},
        edge from parent path={(\tikzparentnode.south) -- ++(0,-1.05cm)
    			-| (\tikzchildnode.north)},
        level 1/.style={sibling distance=7cm,level distance=1.4cm,
    			growth parent anchor=south,nodes=event},
        level 2/.style={sibling distance=7cm},
        level 3/.style={sibling distance=6cm},
        level 4/.style={sibling distance=3cm}
        ]

	\begin{scope}[xshift=0cm,yshift=0cm,very thick,
		node distance=1.6cm,on grid,>=stealth',
		cond/.style={rectangle,draw,fill=cyan!20},
		act/.style={rectangle,draw,fill=green!20},
		flow/.style={rectangle,draw,fill=green!0}]	
	
	\node [flow] (S) [xshift=3cm,yshift=1.2cm]{$\rightarrow$};
	\node [act] (A)  [xshift=-0.0cm, align=left]{\textbf{Land} \\ \small $land := 1$ } edge [<-] (S);
  	\node [flow] (S1)[xshift=1.5cm, yshift=0.1cm]{?} edge [<-] (S);
  	\node [cond] (C1)	[xshift=1.5cm, yshift=-1.6cm, align=left]{\textbf{Accepted} \\\small \sS: $land\_started > 0$ \\ \small \sF: $land\_started < 0$ \\ \small \sR: default } edge [<-] (S1);
  	\node [flow] (B1) [xshift=3cm, yshift=-0.7cm, align=left]{\small subtree for \\ \small exception} edge [<-] (S1);
  	\node [flow] (S2)[xshift=5.2cm, yshift=0.1cm]{?} edge [<-] (S);
  	\node [cond] (C2)	[xshift=5.2cm, yshift=-1.6cm, align=left]{\textbf{Finished} \\\small \sS: $landed > 0$ \\ \small \sF: $landed < 0$ \\ \small \sR: default} edge [<-] (S2);
  	\node [flow] (B2) [xshift=6.6cm, yshift=-0.7cm, align=left]{\small subtree for \\ \small exception} edge [<-] (S2);
	\end{scope}
\end{tikzpicture}
\caption{Multiple responses for a single action example.}
    \label{fig:multipleresponses}
\end{figure}

 Transferring an ability to have \sR ~state to Condition node could be additionally motivated from a mission design point of view. 
Let us assume we need to handle multiple responses for one action (Fig. \ref{fig:multipleresponses}). Now we clearly separate the responses implemented by conditions from call to the external module made action. In contrast, previous works suggested to add extra states for Action nodes \cite{KlocknerStateful}.

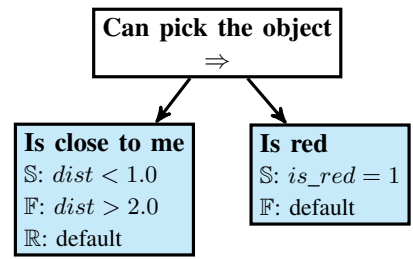
\begin{figure}[h]
    \centering
\begin{tikzpicture}[
        label distance=3mm,
        every label/.style={blue},
        event/.style={rectangle,thick,draw,fill=yellow!20,text width=2cm,
    		text centered,font=\sffamily,anchor=north},
        edge from parent/.style={very thick,draw=black!70},
        edge from parent path={(\tikzparentnode.south) -- ++(0,-1.05cm)
    			-| (\tikzchildnode.north)},
        level 1/.style={sibling distance=7cm,level distance=1.4cm,
    			growth parent anchor=south,nodes=event},
        level 2/.style={sibling distance=7cm},
        level 3/.style={sibling distance=6cm},
        level 4/.style={sibling distance=3cm}
        ]

	\begin{scope}[xshift=0cm,yshift=0cm,very thick,
		node distance=1.6cm,on grid,>=stealth',
		cond/.style={rectangle,draw,fill=cyan!20},
		act/.style={rectangle,draw,fill=green!20},
		flow/.style={rectangle,draw,fill=green!0}]	
	
	\node [flow] (S) [xshift=4.5cm,yshift=1.5cm, align=center]{\textbf{Can pick the object} \\ $\Rightarrow$};
  	\node [cond] (C1)[xshift=3cm, yshift=-0.45cm, align=left]{\textbf{Is close to me} \\ \small \sS: $dist < 1.0$ \\ \small \sF: $dist > 2.0$ \\ \small \sR: default } edge [<-] (S);
  	\node [cond] (C2)[xshift=6cm, yshift=-0.25cm, align=left]{\textbf{Is red} \\ \small \sS: $is\_red = 1$ \\ \small \sF: default } edge [<-] (S);
	\end{scope}
\end{tikzpicture}
\caption{Skipper for treating \sR $ $ state as unknown.}
    \label{fig:skipper}
\end{figure}

\begin{figure}[h]
    \centering
\begin{tikzpicture}[
        label distance=3mm,
        every label/.style={blue},
        event/.style={rectangle,thick,draw,fill=yellow!20,text width=2cm,
    		text centered,font=\sffamily,anchor=north},
        edge from parent/.style={very thick,draw=black!70},
        edge from parent path={(\tikzparentnode.south) -- ++(0,-1.05cm)
    			-| (\tikzchildnode.north)},
        level 1/.style={sibling distance=7cm,level distance=1.4cm,
    			growth parent anchor=south,nodes=event},
        level 2/.style={sibling distance=7cm},
        level 3/.style={sibling distance=6cm},
        level 4/.style={sibling distance=3cm}
        ]

	\begin{scope}[xshift=0cm,yshift=0cm,very thick,
		node distance=1.6cm,on grid,>=stealth',
		cond/.style={rectangle,draw,fill=cyan!20},
		act/.style={rectangle,draw,fill=green!20},
		flow/.style={rectangle,draw,fill=green!0}]	
	\node [flow] (S) [xshift=2.5cm,yshift=1.5cm, align=center]{(0,0) \\ $\Rightarrow$};
  	\node [cond] (C1)[xshift=0cm, yshift=-0.45cm, align=left]{(0,0,0) \\\textbf{Latch condition}: \\ \small \sS: $sub = 1$ and $mem = 1$ \\ \small \sF: $sub = 2$ and $mem = 1$ \\ \small \sR: default } edge [<-] (S);
  	\node [flow] (sel)[xshift=4.5cm, yshift=0.25cm, align=center]{(0,0,1) \\?} edge [<-] (S);
  	\node [flow] (seq)[xshift=3cm, yshift=-1.5cm, align=center]{(0,0,1,0) \\$\rightarrow$} edge [<-] (sel);
  	\node [act] (As)[xshift=5.5cm, yshift=-1.5cm, align=left]{(0,0,1,1) \\ \textbf{Remember \sF}\\ \small $mem := 1$\\ \small $sub := 2$} edge [<-] (sel);
  	\node [flow] (sub)[xshift=0cm, yshift=-3.25cm, align=left]{(0,0,1,0,0) \\ \textbf{Latched subtree}\\ \sR, \sS, \sF} edge [<-] (seq);
  	\node [act] (Af)[xshift=4cm, yshift=-3.38cm, align=left]{(0,0,1,0,1) \\ \textbf{Remember \sS}\\ \small $mem := 1$\\ \small $sub := 1$} edge [<-] (seq);
	\end{scope}
\end{tikzpicture}
\caption{Latch concept implementation with Skipper node. An illustration for nodes order assignment.}
    \label{fig:latch}
\end{figure}
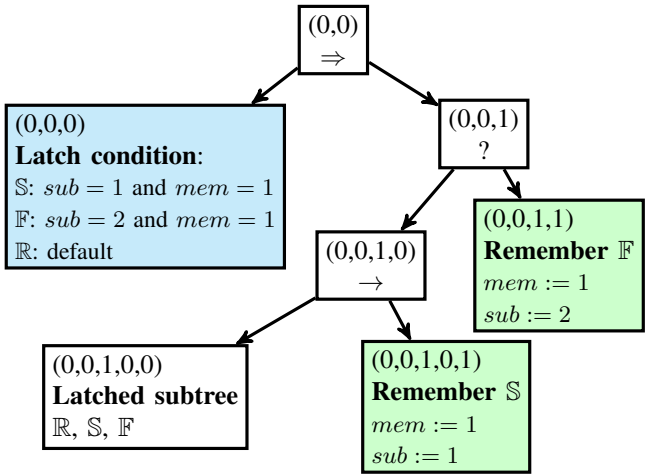

\paragraph{Control nodes} are Sequence, Selector, Parallel, and Skipper. First three are extensively discussed before\cite{SwedishBT}, while Skipper is a new node type. However, it is a ``sibling" of Sequence and Selector in terms of evaluation (see Table \ref{tbl:similarity})
\begin{table}[h]
    \centering
    \caption{Symmetry of three control nodes}
    \label{tbl:similarity}
\begin{tabular}{|l | c c c|}
\hline
 				& Sequence	& Selector 	& Skipper 	\\ \hline
symbol 			& $\rightarrow$ & ? 	& $\Rightarrow$ \\ \hline
Continue on 	& \sS 		& \sF		& \sR		\\ \hline
Return	& \sR, \sF	& \sR, \sS	& \sS, \sF	\\ 
\hline
\end{tabular}
\end{table}
\\
They all tick their children and check the state sequentially from left to right. If the state of the child belongs to the ``Return" subset, control node return the state immediately. In the other case, they continue evaluation and return ``Continue on" state. These three control nodes have similar evaluate functions, see Algorithm \ref{evaluate}.
There are several possible applications of a new control node.  One can treat \sR{} state of a Condition node as \texttt{undefined} or \texttt{unknown} and implement passing the decision making to the next Condition node (Fig. \ref{fig:skipper}).
\\
\paragraph{Latch concept}
another interesting application of Skipper node is an implementation of Latch concept (Fig. \ref{fig:latch}). Latch remembers the first \sS$ $ or \sF$ $ returned state of subtree and do not evaluate the subtree until reseted \cite{KlocknerStateful}.
\\Supplementary Reset node could be implemented just by a single Action node that assigns $mem := 0$ at some place of the tree. 
\\
\section{Asynchronous Tick Propagation}
\begin{figure*}[h]
\centering
\def\svgwidth{\textwidth}
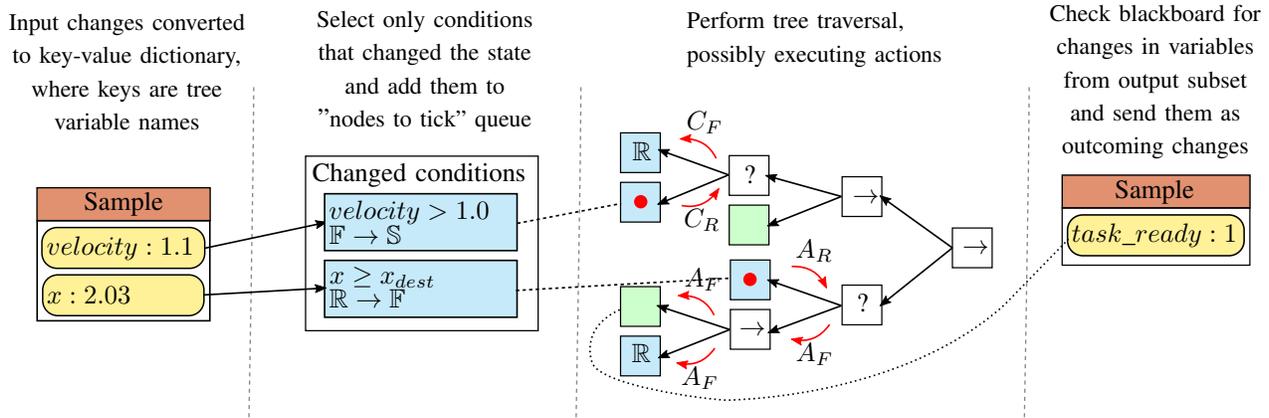

\caption{Pipeline of incoming sample execution. Red arrows show the tick propagation inside the tree.}
\label{fig:pipe}

\end{figure*}

As a mission executor, BT has to acquire input data, respond to this data, modify its own state and sometimes call actions activating changes in corresponding modules of the autonomous system.
Incoming data should first be converted to changes in $\mathcal{V}$. Than, the response to any change to input is done by a \texttt{callback} function. An argument of the function is a dictionary (key - name of variable $v \in \mathcal{V}$, value - new value of $v$). If variables from $Output$ subset were modified through the callback execution, it would be added to the returned dictionary (see also Fig. \ref{fig:pipe}). The callback function consists of four main parts:
\begin{enumerate}
\item Apply the changes to the memory.
\item Re-evaluate conditions, that depends on changed variables. Add conditions which changed their state to the ordered queue $nodes\_to\_tick$.
\item While $nodes\_to\_tick$ queue is not empty, take the \textit{first} node, call its \texttt{tick} method and by returned $state$ and $tick\_type$ \textit{decide} if we add its parent to the queue or not; also add changed conditions, if any.
\item If there are no nodes left to be ticked, return the changes in $Output$ subset of memory. 
\end{enumerate}

\begin{algorithm}
\caption{Callback function}\label{callback}
\begin{algorithmic}[1]
\Function{callback}{\texttt{sample}}\\
\Comment{argument and return type is dict}
	\State $tree.memory.set(sample)$
	\ForAll {$c \in tree.changed\_conditions()$}
		\State $nodes\_to\_tick.insert(c, A_f)$
	\EndFor
 	\While{\textbf{not} $nodes\_to\_tick.empty()$}
		\State $node, tick\_type = nodes\_to\_tick.pop\_front()$
		\State $state, tick\_type = node.tick(tick\_type)$
		\If {$tick\_type \in \{ A_R, C_R \}$}
			\State $nodes\_to\_tick.insert(node.parent)$
		\EndIf
		\ForAll {$c \in tree.changed\_conditions()$}
			\State $nodes\_to\_tick.insert(c, A_f)$
		\EndFor
	\EndWhile\
	\State \textbf{return} $memory.get\_changes(Output)$
\EndFunction
\end{algorithmic}
\end{algorithm}
\paragraph{Tick types}
in synchronous BT tick was always propagated in a top-down or \textit{Fall} manner. In the asynchronous tree we have to define rules for bottom-up or \textit{Rise} propagation. In order to do this, we added a $tick\_type$ parameter for the tick function. Moreover, tick function returns the state of the node and a tick type which should be applied to the parent. For efficient propagation we add another category - ticks might be either \textit{Activating} or \textit{Checking}. The second one is used when we do not need to recursively evaluate subtrees of a control node and could look only on their last state updates. Thus, four tick types are Activating Fall ($A_F$), Activating Rise ($A_R$), Checking Fall ($C_F$), and Checking Rise ($C_R$). $A_F$ represents the propagation manner that is happening in synchronous trees (parent recursively ticks children, we activate actions). $A_R$ is rising of $A_F$ (bottom up, child adds parent to $nodes\_to\_tick$ queue). $A_R$ happens when state changes from Running to Success/Failure implying continuing the paused $A_F$ tick. $C_R$ represents bottom-up propagation of state checking (without activation of any action node). $C_F$ refers to look-up for children state (without further fall). $N_T$ or $\emptyset$ symbol is used when no tick is required.
\par The definition of \texttt{tick} function is the same for each node (Alg. \ref{tick}). The \texttt{evaluate} function is unique for each node type. \texttt{Evaluate} accepts $tick\_type$ argument and returns the resulting state of the node evaluation (Alg. \ref{evaluate}). Definition of both \texttt{tick} and \texttt{evaluate} functions is discussed later after explaining tick propagation rules.
\paragraph{Order of nodes in the tree} undefined order of bottom-up propagation of multiple conditions' changes could lead to different output results. Thus, we have to clarify the order of nodes in $nodes\_to\_tick$ priority queue. One needs to ensure that the parent would not be evaluated before the child and children would be evaluated from left to right. These rules are exactly the description of Kleene-Brouwer order. To compare, each node holds an order, which is an array of integers. The root node has an order containing one number $\{0\}$. To make a child order, we append its position in children list to the order array of the parent. See also a behavior tree example with node order (Fig. \ref{fig:latch}).
\paragraph{Propagation rules}
the propagation rules could be clearly defined by two tables for each type of a node. First table -- return tick table (see Table \ref{tbl:return}) -- is the same for all node types and illustrates which tick type would be applied to the parent node in the case of bottom-up propagation. The returned tick type depends only on states before and after node evaluation.
\begin{table}[h]
    \centering
    \caption{Return table}
    \label{tbl:return}
\begin{tabular}{| c  l | c  c  c|}
\hline
& &  \multicolumn{3}{c|}{State after tick} \\
  	& & \sR	& \sS 	& \sF 	\\ \hline
State  & \sR & \nt   & $A_R$	& $A_R$	\\
before & \sS	& \nt	& \nt	& $C_R$	\\
tick & \sF	& \nt 	& $C_R$	& \nt	\\
\hline
\end{tabular}
\end{table}
Hence the diagonal elements of the table are \nt, meaning there would be no bottom-up propagation in case the state of the node was not changed. The same holds true if the state changed to \sR$ $ from either \sS$ $ or \sF. This change could be treated as activating from the bottom to the top. We forbid such changes because task activation is naturally triggered from the top to the bottom. However, altering elements in a table is a flexible way to change rules of tick propagation. Success to Failure changes are $C_R$ because they do not imply any local task resumption. Finally, \sR $ $ to \sS $ $ or \sF $ $ changes are $A_R$. 
\begin{table}[h]
\caption{Call table for sequential control nodes}    
    \centering
\begin{tabular}{| c  l | l  c  c  r|}
\hline
		&		& \multicolumn{4}{c|}{tick type argument}\\
  		&		& $A_F$	& $A_R$ & $C_F$	& $C_R$	\\ \hline
State	& \sR 	& $A_F$	& $A_F$	& \nt	& $C_F$	\\
before	& \sS	& $A_F$	& \nt	& \nt	& $C_F$	\\
tick	& \sF	& $A_F$ & \nt	& \nt	& $C_F$ \\
\hline
\end{tabular}
    \label{tbl:callcontol}
\end{table}

\par Table \ref{tbl:callcontol} -- call table -- illustrates how the control node should evaluate the children. The tick type passed as argument to \textit{evaluate} function depends on a current state and a tick type passed as argument to \textit{tick} function. Different control nodes might have different evaluation tick tables. Let's thoroughly describe the content of the evaluation tick table for Sequence. $A_F$ column represents the activation process. All the nodes are recursively evaluated, encountered actions are called. The rest is specific for asynchronous BT. The second column in \sS $ $ and  \sF $ $ rows shows the ban for bottom-up activation propagation. Last 2 columns mean that for $\msS \leftrightarrow \msF$ transition tree does not resume any activation process. 
As for the Parallel node we can change $A_F$ to $C_F$ in the $A_R$ column. All subtrees are activated through the initial $A_F$ tick, therefore, the node does not need to activate them later.

\paragraph{Tick and evaluate functions}
tick function is the same for every node type (see the definition in Alg. \ref{tick}). Its main purpose is to obey propagation rules defined above and call the evaluate function with proper \texttt{tick\_type} argument. 

\begin{algorithm}
\caption{Evaluate function for Sequential}\label{evaluate}
\begin{algorithmic}[1]
\Function{node.evaluate}{$tick\_type$}\\
\Comment{returns $state$}
	\If {$tick\_type = \mnt$}
		\State \textbf{return} $node.state()$
	\EndIf
	\ForAll {$child \in node.children()$}
		\State $state, tick\_type = child.tick(tick\_type)$
		\If {$state \in node.type.Return}$ \Comment{see Table \ref{tbl:similarity}}
			\State \textbf{return} $state$
		\EndIf
	\EndFor
	\State \textbf{return} $node.type.ContinueOn$ \Comment{see Table \ref{tbl:similarity}}
\EndFunction
\end{algorithmic}
\end{algorithm}

\begin{algorithm}
\caption{Tick function}\label{tick}
\begin{algorithmic}[1]
\Function{node.tick}{$tick\_type$}\\
\Comment{returns $state, tick\_type$}
	\State $old\_state = node.state()$
	\State $child\_tick\_type = call\_table[old\_state][tick\_type]$
	\State $node.set\_state(node.evaluate(child\_tick\_type))$ 	
	\State \textbf{return} $return\_table[old\_state][node.state()]$
\EndFunction
\end{algorithmic}
\end{algorithm}

\paragraph{Synchronous tree inside asynchronous}
Synchronous BT could be implemented as a subtree inside asynchronous having a periodical time event (see Fig. \ref{fig:synctree})  

\begin{figure}[h]
    \centering
\begin{tikzpicture}[
        label distance=3mm,
        every label/.style={blue},
        event/.style={rectangle,thick,draw,fill=yellow!20,text width=2cm,
    		text centered,font=\sffamily,anchor=north},
        edge from parent/.style={very thick,draw=black!70},
        edge from parent path={(\tikzparentnode.south) -- ++(0,-1.05cm)
    			-| (\tikzchildnode.north)},
        level 1/.style={sibling distance=7cm,level distance=1.4cm,
    			growth parent anchor=south,nodes=event},
        level 2/.style={sibling distance=7cm},
        level 3/.style={sibling distance=6cm},
        level 4/.style={sibling distance=3cm}
        ]

	\begin{scope}[xshift=0cm,yshift=0cm,very thick,
		node distance=1.6cm,on grid,>=stealth',
		cond/.style={rectangle,draw,fill=cyan!20},
		act/.style={rectangle,draw,fill=green!20},
		flow/.style={rectangle,draw,fill=green!0}]	
		
	\node [flow] (R) [xshift=4.9cm,yshift=5.7cm, align=center]{$\rightarrow$}; 
	\node [cond] (trig) [xshift=1.6cm,yshift=4.6cm, align=left]{\textbf{Wait for a next tick}  \\ \small \sS: $time < t_{prev} + \Delta t $ \\ \small \sR: default} edge [<-] (R);
	\node [act] (tset) [xshift=4.9cm,yshift=4.6cm, align=left]{\textbf{Set next tick time} \\ $t_{prev} := time$ } edge[<-] (R);
	\node [flow] (subt) [xshift=7.6cm,yshift=4.73cm, align=left]{\textbf{Synchronous} \\ \textbf{subtree}} edge [<-] (R);
  	\end{scope}
\end{tikzpicture}

    \caption{Synchronous subtree implementation.}
    \label{fig:synctree}
\end{figure}
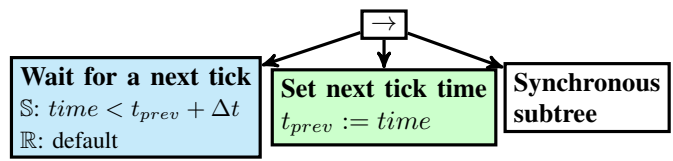

\begin{figure*}[h]
    \centering
\begin{tikzpicture}[
        label distance=3mm,
        every label/.style={blue},
        event/.style={rectangle,thick,draw,fill=yellow!20,text width=2cm,
    		text centered,font=\sffamily,anchor=north},
        edge from parent/.style={very thick,draw=black!70},
        edge from parent path={(\tikzparentnode.south) --  (\tikzchildnode.north)},
        level 1/.style={sibling distance=7cm,level distance=1.4cm,
    			growth parent anchor=south,nodes=event},
        level 2/.style={sibling distance=7cm},
        level 3/.style={sibling distance=6cm},
        level 4/.style={sibling distance=3cm}
        ]

	\begin{scope}[xshift=0cm,yshift=0cm,very thick,
		node distance=1.6cm,on grid,>=stealth',
		cond/.style={rectangle,draw,fill=cyan!20},
		act/.style={rectangle,draw,fill=green!20},
		flow/.style={rectangle,draw,fill=green!0}]	
	\node [flow] (R) [xshift=7cm,yshift=5.5cm, align=center]{\textbf{sync\_tree} \\ $\rightarrow$}; 
	\node [cond] (start) [xshift=0cm,yshift=4.9cm, align=left]{\textbf{Start synchronization} \\ \sS: $trigger\_sync = 1$ or\\ $\exists i$ $hash\_set\_\{i\} = 1$ \\ \sR: default } edge [<-] (R);
	\node [flow] (lsent) [xshift=2.7cm,yshift=4.5cm, align=center]{$\widehat{Latch}$ } edge[<-] (R);
	\node [act] (hsend) [xshift=2.7cm,yshift=3.cm, align=left]{\textbf{Send hash} \\ $hash\_1 := tree.hash()$ \\ $hash\_set\_1 := 1$ \\ $time\_start = time$} edge [<-] (lsent);
	\node [flow] (na) [xshift=7cm,yshift=3cm, align=center]{?} edge [<-] (R);
	\node [flow] (lrec) [xshift=3cm,yshift=1.5cm, align=center]{$\widehat{Latch}$ } edge[<-] (na);
	\node [flow] (hsrss) [xshift=3cm,yshift=0.7cm, align=center]{$\Rightarrow$} edge [<-] (lrec);
	\node [flow] (hss) [xshift=0cm,yshift=0.3cm, align=center]{\textbf{All hashes recieved} \\ $\rightarrow$} edge[<-] (hsrss);
	\node [cond] (set1) [xshift=0cm,yshift=-1.1cm, align=left]{$\msS: hash\_set\_1 = 1$ \\ \sR: default} edge [<-] (hss);
	\node [cond] (set2) [xshift=0.7cm,yshift=-2.1cm, align=left]{$\msS: hash\_set\_2 = 1$ \\ \sR: default} edge [<-] (hss);
	\node [cond] (set3) [xshift=1.4cm,yshift=-3.1cm, align=left]{$\msS: hash\_set\_3 = 1$ \\ \sR: default} edge [<-] (hss);
	\node [cond] (timeout) [xshift=4cm,yshift=-1.8cm, align=center]{\textbf{Timeout}\\ $time - time\_start$\\ $> max\_delay$} edge [<-] (hsrss); 
	\node [flow] (chm1) [xshift=7.2cm,yshift=1.9cm, align=center]{\textbf{Choose a new master} \\?} edge [<-] (na);
	\node [flow] (sm1) [xshift=6.5cm,yshift=0.9cm, align=center]{ $\rightarrow$} edge [<-] (chm1);
	\node [cond] (m1c) [xshift=6cm,yshift=-0.35cm, align=left]{\textbf{if 1$^\text{st}$ alive} \\$\msS: hash\_set\_1 = 1$ \\ \sF: default} edge [<-] (sm1);
	\node [act] (m1a) [xshift=7cm,yshift=-1.4cm, align=left]{$master := 1$} edge [<-] (sm1);
	\node [flow] (chm2) [xshift=9cm,yshift=-0.5cm, align=center]{?} edge [<-] (chm1);
	\node [flow] (sm2) [xshift=9cm,yshift=-1.5cm, align=center]{$\rightarrow$} edge [<-] (chm2);
	\node [cond] (m2a) [xshift=7.5cm,yshift=-2.75cm, align=left]{\textbf{if 2$^\text{nd}$ alive} \\$\msS: hash\_set\_2 = 1$ \\ \sF: default} edge [<-] (sm2);
	\node [act] (m2a) [xshift=10.4cm,yshift=-2.3cm, align=left]{$master := 2$} edge [<-] (sm2);
	\node [act] (m3a) [xshift=12cm,yshift=-1.5cm, align=left]{$master := 3$} edge [<-] (chm2);
	\node [flow] (ifalive) [xshift=11.2cm,yshift=4.3cm, align=center]{?} edge[<-] (R); 
	\node [cond] (ifal) [xshift=9.6cm,yshift=3.1cm, align=left]{\textbf{\sS $ $ if all alive have } \\ \textbf{same $hash\_\{i\}$ values} \\ \sF: default} edge[<-] (ifalive);
	\node [flow] (ifmaster) [xshift=12.3cm,yshift=3.3cm, align=center]{?} edge[<-] (ifalive);
	\node [flow] (mas) [xshift=11.5cm,yshift=1.5cm, align=center]{$\rightarrow$ } edge[<-] (ifmaster);
	\node [cond] (ifm) [xshift=11cm,yshift=0.2cm, align=center]{\textbf{If I am master FCC} \\ \sS: $me == master$ \\ \sF: default} edge[<-] (mas);
	\node [act] (ssend) [xshift=14.2cm,yshift=0.5cm, align=center]{$send\_vars := 1$} edge[<-] (mas);
	\node [cond] (ifm) [xshift=14cm,yshift=1.8cm, align=center]{\sS: $recieved\_vars == 1$ \\ \sF: default} edge[<-] (ifmaster);
	\node [act] (cleanup) [xshift=14.3cm,yshift=4.5cm, align=left]{\textbf{Finish} \\ $trigger\_sync := 0$ \\ and clean up all \\ neccessary  variables} edge[<-] (R);
  	\end{scope}
\end{tikzpicture}

    \caption{Complementary behavior tree for 3 FCC unit synchronization logic.}
    \label{fig:syncmission}
\end{figure*}
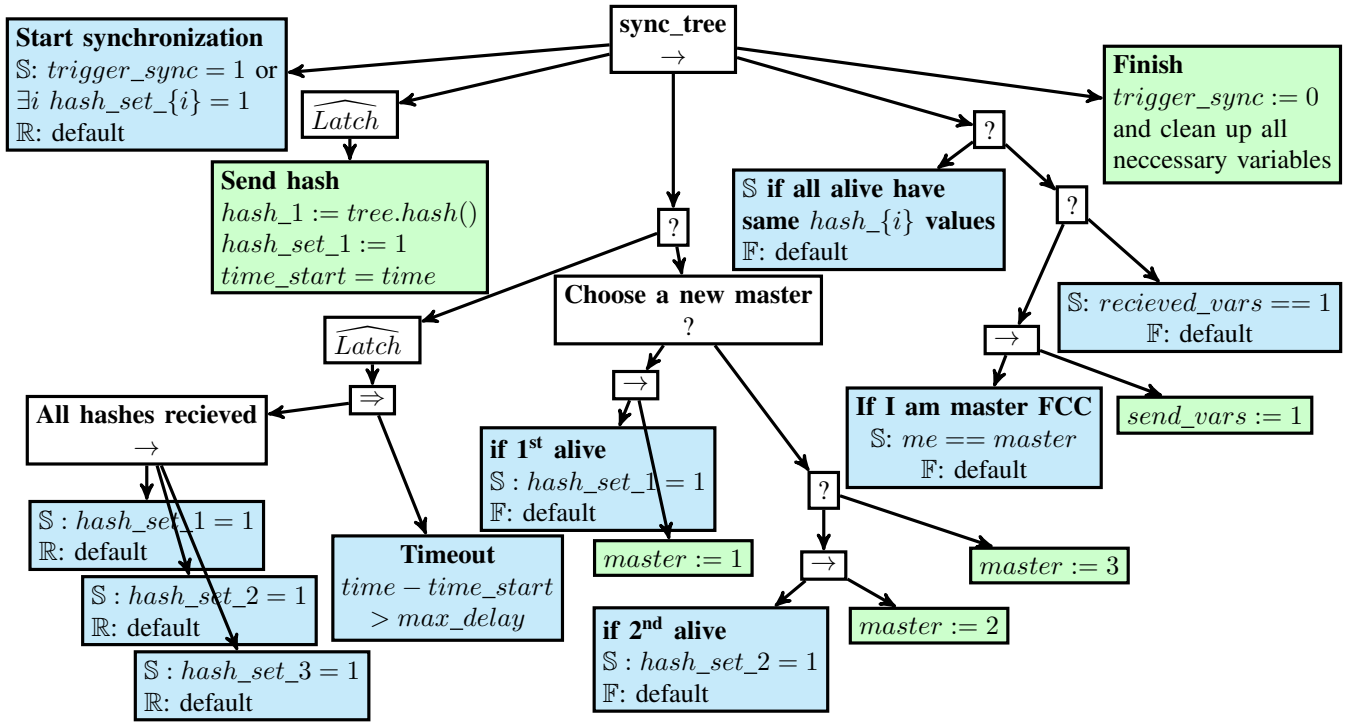

\section{Execution With Redundancy on Flight Control Computers}
We need to synchronize the states of FCC mission execution systems.
In our system, any changes are transferred to the outer scope only after calling \texttt{callback} function. Assume we started the same trees on all FCCs. Then, result of \texttt{callback} depends only on variable values $var(\mathcal{V})$ and node states $state(\mathcal{N}) \subset \mathcal{V}$. Thus, if all three FCCs have the same $\mathcal{V}$ before callback execution, they would have the same output. That means we need to be sure in variable values constistency before each callback. In fact, we have to synchronize variables only if some condition has changed. If no condition changed, there would be no tick propagation started and no output changes produced. Condition change events usually lead to the execution of some action and, thus, are much more rare than incoming sample events. The developed memory and asynchrony features of BT helped to drastically reduce the amount of synchronization events.
\par This logic was implemented by modifying \texttt{callback} function and adding an extra independent BT \texttt{sync\_tree} for synchronization.  Fig. \ref{fig:syncmission} shows the concept of this complementary BT. It consists of 5 main branches:
\begin{enumerate}
\item Condition that activates further execution.
\item Latched Action node that sends a hash of $\mathcal{V}$ to other FCCs.
\item Subtree that waits for the hashes from other FCCs and chooses a new master FCC if the current master is not responding.
\item Subtree that sends changes in $\mathcal{V}$ from the master FCC to slaves if their hashes differ.
\item Action that returns all variables to their initial values.
\end{enumerate}

\begin{algorithm}

\caption{Callback function with synchronization}\label{callbackwsync}
\begin{algorithmic}[1]
\Function{callbackWithSync}{\texttt{sample}}\\
\Comment{sample and return value are dicts}
	\If {\textbf{not} $is\_sync\_sample(sample)$}
		\State $tree.memory.set(sample)$
		\If { $tree.changed\_conditions()$}
			\State $sample := \{"trigger\_sync": 1\}$
		\EndIf
	\EndIf
	\If {$is\_sync\_sample(sample)$}
		\State $sync\_res = sync\_tree.callback(sample)$
		\If {$"sync\_ended" \in sync\_res$}
			\State \textbf{return} $sync\_res + tree.callback(dict())$
		\Else
			\State \textbf{return} $sync\_res$
		\EndIf
	\EndIf
	\State \textbf{return} dict()
\EndFunction
\end{algorithmic}
\end{algorithm}

\par Real synchronization procedure and corresponding behavior tree might be even more complicated (e.g. timeout conditions for any communication with other FCCs). 

\section{Implementation and Tests}
\begin{figure}[ht]
\centering
\includegraphics[width=0.48\textwidth]{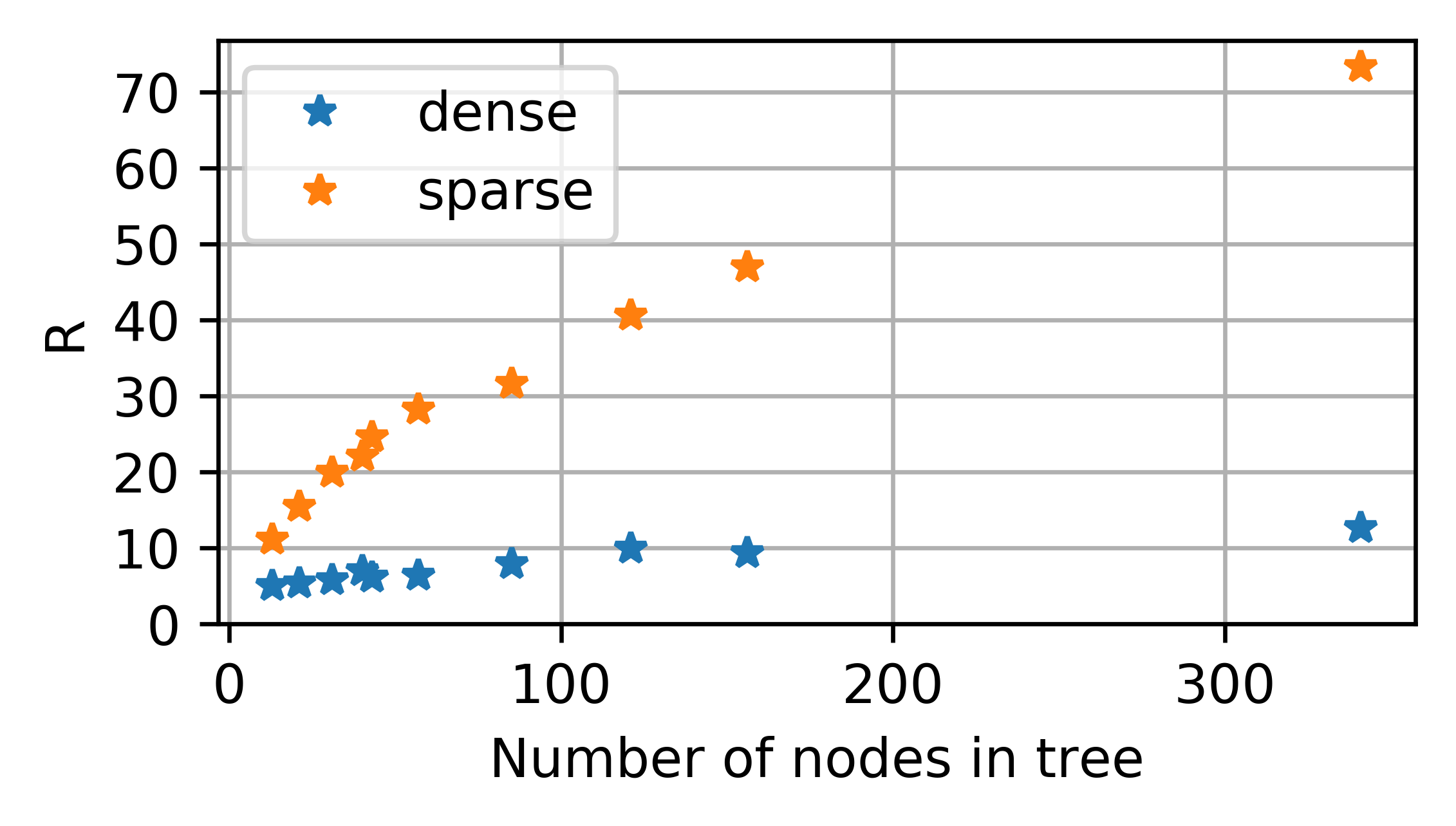}
\caption{Computational costs evaluation for classical BT and asynchronous tick propagation.}
\label{fig:eval}
\end{figure}

\paragraph{C++ framework}
we developed a C++ framework with the implementation of described Asynchronous Behavior Tree with Memory (ABTM) concept. It includes tree construction and execution. For leaf nodes one can use C++ functions or construct nodes from simple expressions (such as $velocity == 0$ or $x = 1$). The framework was developed with thread safety taken into account.
\paragraph{Real-time simulation}
in order to verify our concepts, we also implemented synchronization procedure following the guideline from Section IV. 
We connected our framework to the real-time HAPS mission simulation. It imitates the behavior of all other aircraft modules, including high-frequency data flow from positioning and all control modules. To prove our system working we started three independent mission executors. Only one current master executor was allowed to send output to other modules. All FCCc are shown to maintain the same state in long-time runs. Another test was a crash imitation by simultaneously stopping any of the executors at random time. In this case, the remaining one or two executors chose a new master and the mission continued correctly. 
\paragraph{Efficiency evaluation}
we compared the computation times for asynchronous execution (Section III) and classical top down tree traversal (applying $A_R$ to the root). To eliminate the difference between certain BT structures we created 200 random tree structures with height in the range of $3..5$. 
Number of each control node children was in the range of $3..7$. We averaged the results for different complexities of conditions. ABTM was proven to be faster in all of the cases. We plotted a ratio between classical and asynchronous time execution (Fig. \ref{fig:eval}).
\begin{equation}
R = \frac{time_{classical~BT}}{time_{ABTM}}
\end{equation}
\par Dense case illustrates the situation when every incoming sample changes at least one condition and starts tree traversal for ABTM. Sparse case is the opposite to the dense -- every sample does not trigger any condition. ABTM outperforms classical approach in all cases with ratio $R \in [10, 70]$ for trees with $300$ nodes. This feature speeds up the continous time simulation. \par In addition, ABTM reduces the amount of synchronization call in case of redundant execution (Section IV). However, the difference highly varies with application. If the robot executes an action every $3$ seconds and classical BT runs with $20~Hz$ update rate, then the amount of synchronization calls would be reduced by $3 \cdot 20 = 60$ times.

\section{Conclusions}
We formulated a new Asynchronous behavior tree with memory approach. It includes a description of the memory layer, rules of tick propagation, control and leaf nodes definition. Overall, the developed BT functionality is richer and more flexible than previous definitions. Lower computational costs are beneficial for continous time simulations. In addition, asynchronous approach helped us decrease the amount of synchronization calls in execution on redundant FCCs.
\par
Our approach also allowed to make a standalone C++ library. An ability for automated testing the BT logic outside the robot environment in a simple simulation mode (e.g. applying necessary input changes and checking the outputs) is also a benefit.
\par Our changes in BT semantics, especially memory layer, simplified the development of the synchronization algorithm for multiple FCCs. We demonstrated the possibility to use our framework on robotic systems with redundancy on FCC units and tested it in the simulation. 
\par
Our future plans include flight experiments of unmanned HAPS missions using a developed framework with triple redundancy support. 

\bibliography{iros2019}
\bibliographystyle{ieeetr}
\end{document}